\documentclass{article} % For LaTeX2e
\usepackage{iclr2026_conference,times}

\usepackage{amsmath,amsfonts,bm}

\def\eqref#1{equation~\ref{#1}}
\def\1{\bm{1}}

\DeclareMathAlphabet{\mathsfit}{\encodingdefault}{\sfdefault}{m}{sl}
\SetMathAlphabet{\mathsfit}{bold}{\encodingdefault}{\sfdefault}{bx}{n}

\usepackage{hyperref}
\usepackage{url}

\title{\MODEL: Automatic Benchmark Evolution via Multi-Model Competitive Evaluation}

\author{Qin Liu$^{1}$ \;
    Jacob Dineen$^{2}$ \;
    Yuxi Huang$^{2}$ \;
    Sheng Zhang$^{3}$ \;
    Hoifung Poon$^{3}$ \; \\
    \textbf{Ben Zhou}$^{2}$ \;
    \textbf{Muhao Chen}$^{1}$ \\
$^{1}$University of California, Davis \;
$^{2}$Arizona State University \;
$^{3}$ Microsoft Research \\
\texttt{qinli@ucdavis.edu} \; \texttt{\{jdineen, yhuan504\}@asu.edu} \; \\
\texttt{\{shezhan, hoifung\}@microsoft.com}\\
\texttt{xzhou202@asu.edu} \; \texttt{muhchen@ucdavis.edu}
}

\usepackage{graphicx}
\usepackage{algorithm}
\usepackage{algpseudocode}
\usepackage{booktabs}
\usepackage{multirow}
\usepackage[table]{xcolor}
\usepackage{pifont}
\usepackage{xcolor}

\usepackage{listings}
\usepackage{cleveref}
\crefformat{section}{\S#2#1#3}
\crefformat{subsection}{\S#2#1#3}
\crefformat{subsubsection}{\S#2#1#3}
\crefrangeformat{section}{\S#3#1#4 to~\S#5#2#6}
\crefmultiformat{section}{\S#2#1#3}{ and~\S#2#1#3}{, #2#1#3}{ and~#2#1#3}
\Crefformat{figure}{#2Fig.~#1#3}
\Crefmultiformat{figure}{Figs.~#2#1#3}{ and~#2#1#3}{, #2#1#3}{ and~#2#1#3}
\Crefformat{table}{#2Tab.~#1#3}
\Crefmultiformat{table}{Tabs.~#2#1#3}{ and~#2#1#3}{, #2#1#3}{ and~#2#1#3}
\Crefformat{appendix}{#2Appx.~\S#1#3}
\crefformat{algorithm}{Alg.~#2#1#3}
\Crefformat{equation}{#2Eq.~#1#3}

\newcommand{\stitle}[1]{\vspace{1ex} \noindent{\bf #1.}}

\usepackage{xspace}
\newcommand{\MODEL}{\mbox{\textsc{ArenaBencher}}\xspace}

\iclrfinalcopy % Uncomment for camera-ready version, but NOT for submission.
\begin{document}

\maketitle

\begin{abstract}
Benchmarks are central to measuring the capabilities of large language models and guiding model development, yet widespread data leakage from pretraining corpora undermines their validity. Models can match memorized content rather than demonstrate true generalization, which inflates scores, distorts cross-model comparisons, and misrepresents progress. We introduce \MODEL, a model-agnostic framework for automatic benchmark evolution that updates test cases while preserving comparability. Given an existing benchmark and a diverse pool of models to be evaluated, \MODEL infers the core ability of each test case, generates candidate question–answer pairs that preserve the original objective, verifies correctness and intent with an LLM as a judge, and aggregates feedback from multiple models to select candidates that expose shared weaknesses. The process runs iteratively with in-context demonstrations that steer generation toward more challenging and diagnostic cases. We apply \MODEL to math problem solving, commonsense reasoning, and safety domains and show that it produces verified, diverse, and fair updates that uncover new failure modes, increase difficulty while preserving test objective alignment, and improve model separability. The framework provides a scalable path to continuously evolve benchmarks in step with the rapid progress of foundation models.
\end{abstract}

\section{Introduction}

%{\color{blue} \textbf{change all queries, items etc. to test cases.}}

Benchmarks %have become a critical tool for evaluating language model
are indispensable for assessing large language models (LLM)
capabilities and steering model development \citep{gsm8k,winogrande,csqa,mmlu,bigbench,helm}. %However, concerns have emerged that many widely used benchmarks have been partially or entirely included in the training data of major language models. 
Yet growing evidence that widely used benchmarks are partially or fully present in pretraining corpora of models poses a fundamental validity threat: models can exploit memorized content rather than demonstrating true generalization \citep{wu2025reasoningmemorizationunreliableresults,liang2025largelanguagemodelcheat,xu2024benchmarkingbenchmarkleakagelarge,dong-etal-2024-generalization,balloccu-etal-2024-leak,jiang2024investigatingdatacontaminationpretraining}.
This pervasive data leakage fundamentally undermines the reliability of evaluation, %which can cause models to appear more capable than they truly are by leveraging memorized rather than generalized knowledge. The result is inflated performance, unfair comparisons, and misleading progress.
causing inflated reporting scores, distorted cross-model comparisons, and misrepresented progress of development.
These risks motivate evaluation methods that continually refresh and harden benchmarks against leakage while preserving comparability over time \citep{li2024autobencher,wang2025evolmathevalevolvablebenchmarksmathematical}.

%Two common strategies are to augment or regenerate benchmark items through automatic paraphrasing or adversarial perturbation.
Prior efforts typically augment or modify test cases in benchmarks via paraphrasing or adversarial perturbations to raise difficulty and reduce overlap while preserving task intent.
%These methods aim to increase difficulty or reduce data overlap while preserving the original task intent. 
In mathematical reasoning, for example, works often perturb surface details such as numerical values or swap concepts within a confined symbolic space \citep{yang-etal-2025-evaluating,abedin2025arithmattack,gsm_symbolic,math_perturb}. These tweaks raise local difficulty but rarely generalize across task types or domains. 
Other methods use gradient-based adversarial techniques to %probe model robustness by finding perturbations that maximize the loss with respect to a particular model. 
maximize loss for a particular model, yielding test case variants tuned to that model's weaknesses \citep{Liu2024AutoDAN,mo2025redcoderautomatedmultiturnred}.
%However, these approaches inherently optimize against a single model, leading to benchmark variants with strong model-specific biases that may exhibit poor transferability across different systems. 
Such single-model optimization introduces model-specific bias such that test cases that stump one system can be trivial for others, producing evaluation artifacts, unstable rankings, and opaque cross-model comparisons.
%Items that are difficult for one model may remain trivial for others, leading to evaluation artifacts, instability in model rankings, and a lack of interpretability in cross-model comparisons. 
These shortcomings call for benchmark construction methods that are generalizable, model-agnostic, and yield challenges that are both challenging and fair across diverse language models.
We summarize the existing techniques and compare them with our proposed framework in \Cref{tab:benchmark-comparison}.

To address these limitations, we propose \MODEL, a model-agnostic framework for \emph{automatic benchmark evolution} that prioritizes cross-model fairness, model separability, and task alignment. 
Given an existing benchmark and 
%The framework takes as input an existing benchmark in need of updating and 
a pool of diverse language models, \MODEL first infers the core \emph{``ability''} targeted by each test case. For example, a test case in a math reasoning benchmark may test multi-step arithmetic such as chained addition or division, while a test case from a safety benchmark may assess the model’s ability to detect and reject harmful actions described indirectly. 
Based on the extracted ability, \MODEL generates candidate query-label pairs that preserve the original task objective while introducing controlled variation. 
Each candidate is first verified by an LLM-as-a-judge to ensure label correctness and alignment with the intended ability.
To assess candidate effectiveness, \MODEL probes a random subset of models and uses score candidates using their collective feedback, e.g., aggregated loss values or behavioral failures. Candidates that \emph{consistently degrade performance across the sampled models} are prioritized, as they are more likely to reflect shared failure patterns and generalizable weaknesses across different models. This multi-model evaluation mitigates individual model biases, reduces the risk of overfitting, and encourages the discovery of test cases that are broadly challenging across models. To further boost the quality of benchmark updates, \MODEL performs iterative refinement for test cases. After each round of candidate generation and evaluation, the strongest candidates are retained as in-context demonstrations to guide subsequent generations. This process allows \MODEL to progressively steer generation toward more challenging and targeted cases, amplifying common failure signals while preserving alignment with the original task intent.

\newcommand{\cmark}{\color{teal}{\ding{51}}} 
\newcommand{\xmark}{\textcolor{red}{\ding{55}}}   

\begin{table}[t]
\centering
\small
\caption{Comparison of previously proposed benchmark update frameworks. \MODEL supports multi-model, multi-objective, and domain-general benchmark evolution.}
\vspace{0.4em}
\label{tab:benchmark-comparison}
\resizebox{\textwidth}{!}{%
\begin{tabular}{l@{\hskip 1em}lccccc}
\toprule
\textbf{ } & \textbf{Domain} & \textbf{Fairness} & \textbf{Difficulty} & \textbf{Separability} & \textbf{Alignment} & \textbf{Generality} \\
\midrule
MATH-Perturb \citep{math_perturb}        & Math         & \cmark & \cmark & \xmark & \xmark & \xmark \\
\textsc{AR-Checker} \citep{hou2025automatic}      & Math         & \xmark & \cmark & \xmark & \xmark & \xmark \\
AutoBencher \citep{li2024autobencher}         & Multi & \cmark & \cmark & \cmark & \xmark & \cmark \\
AutoDAN \citep{Liu2024AutoDAN}             & Safety       & \xmark & \cmark & \xmark & \cmark & \xmark \\
\midrule
\textbf{\MODEL (Ours)} & Multi & \cmark & \cmark & \cmark & \cmark & \cmark \\
\bottomrule
\end{tabular}
}
\end{table}

We consider four desiderata to evaluate the quality of benchmark updates. (1) \emph{Separability}: the updated test case should induce more variance in model performance, revealing clearer differentiation across systems. (2) \emph{Fairness}: any performance drop should be comparably distributed, avoiding model-targeted artifacts. (3) \emph{Alignment}: the updated test case should preserve the original task objective and core ability so the evaluation intent remains unchanged.
(4) \emph{Difficulty}: the updated test case should be more challenging for the model pool and expose additional model failure modes.
%Under these criteria, a high-quality update is one that is better at exposing performance gaps, remains faithful to the original evaluation intent, and does not introduce systematic bias.
An update that satisfies these criteria exposes genuine performance gaps while remaining faithful to the task and free of systematic bias.

Overall, this work makes threefold contributions. First, we introduce \MODEL, a benchmark-evolution framework that aggregates feedback from a diverse set of language models, mitigating the bias and overfitting associated with single-model evaluation. Second, we design an ability-aware, failure-sensitive update mechanism that %identifies the core skill targeted by each query and selects new candidates that consistently degrade performance across models.
infers the core skill of each test case and selects candidates that consistently depress performance across models.
Third, we develop an iterative refinement strategy in which strong candidates are reused as in-context demonstrations to steer future generations toward increasingly challenging and diagnostic test cases. 
%We demonstrate the effectiveness of \MODEL across both capability and safety evaluation settings, showing that it produces benchmark updates that are more difficult, more discriminative, and more reliable for comparative assessment.
Evaluated on both capability and safety settings, \MODEL yields updates that are more difficult, more discriminative, and more reliable for comparative assessment.

\section{Related Work}

\stitle{Benchmarking Language Models}
Benchmarks serve as the primary instruments for assessing the evolving capabilities of large language models. Early efforts focused on domain-specific tasks such as arithmetic reasoning and commonsense inference, exemplified by GSM8K \citep{gsm8k} for grade-school math and Winogrande \citep{winogrande} for pronoun resolution. More comprehensive benchmarks aggregate a diverse set of tasks to provide holistic evaluations, including MMLU \citep{mmlu} for academic knowledge, BIG-bench \citep{bigbench} for broad capability probing, and HELM \citep{helm} and the Open LLM Leaderboard \citep{open-llm-leaderboard} for standardized reporting across models. Recent work has also sought to improve efficiency by subsampling (BIG-bench Lite) or compressing large benchmarks while preserving model ranking patterns \citep{li2024active,perlitz2023efficient}. To further scale evaluation, LLM-as-a-Judge methods employ models to automatically grade responses, reducing annotation costs and enabling continuous assessment at scale \citep{zheng2023judging,lee2023rlaif,bai2022constitutional,
gu2024survey}. Despite their broad coverage, these benchmarks remain static datasets and are vulnerable to data contamination, where test items overlap with training corpora \citep{benchmark_contamination_survey,dekoninck2024evading,choicontaminated}. As a result, model scores may reflect memorization rather than genuine generalization. Moreover, static test sets often fail to adapt to frontier models, leading to saturated performance and reduced discriminative power, prompting recent efforts to develop manual and automatic update mechanisms \citep{benchmark_contamination_advancements,jain2024livecodebench, white2024livebench,fan2023nphardeval,ying2024automating,li2024autobencher}. These limitations motivate \MODEL, which aims to evolve benchmarks dynamically so that they remain challenging, fair, and aligned with their original evaluation intent.

\begin{figure}[t]
   \centering 
   \includegraphics[width=0.95\textwidth]{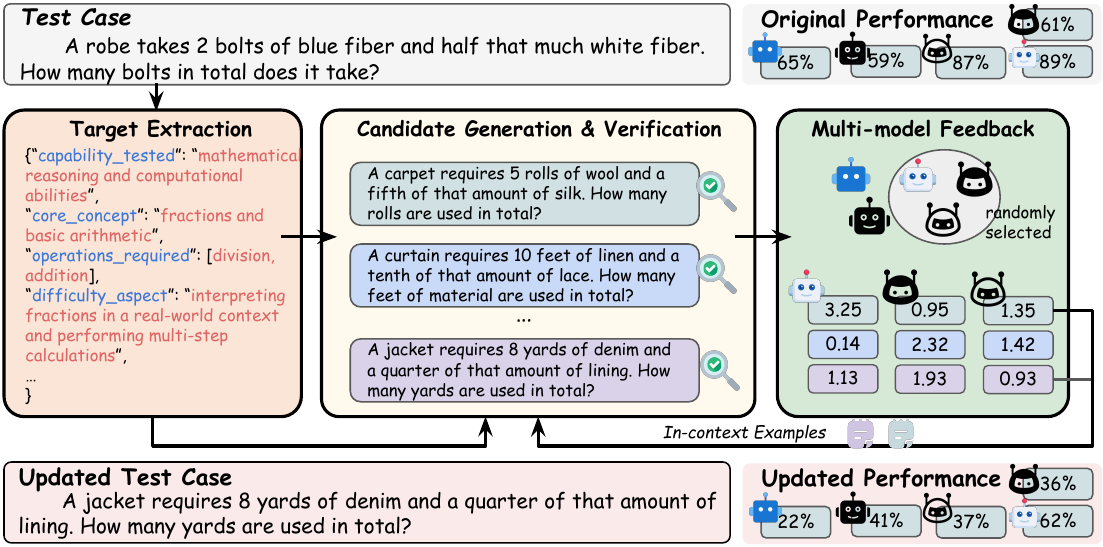}
   \caption{Overview of \MODEL on a math reasoning example. Starting from an original test case, the system extracts a structured target objective that specifies multiple rubrics. Conditioned on this objective and in-context demonstrations of strong candidates evaluated by multi-model feedback, the generator iteratively proposes multiple candidate queries and answers, and an independent LLM judge verifies correctness and test target alignment.}
    \label{fig:main}
\end{figure}

\stitle{Benchmark Augmentation}
A complementary line of work aims to augment existing benchmarks by generating variants of test instances that probe model robustness \citep{hong2024evaluating}. This strategy has been especially prominent in mathematical reasoning, where benchmarks such as GSM8K, MATH \citep{math-500}, and AIME \citep{aime} have been systematically perturbed to reveal brittleness. For example, simple perturbations that replace numerical values with variables or inject punctuation marks into math word problems have been shown to significantly reduce model accuracy \citep{yang-etal-2025-evaluating,abedin2025arithmattack,gsm_symbolic}. More sophisticated approaches construct semantically equivalent but syntactically altered problems. For instance, MATH-Perturb \citep{math_perturb} categorizes transformations into ``simple'' and ``hard'' perturbations, showing that leading models suffer large drops on hard perturbations where the reasoning steps fundamentally change \citep{yu2025benchmarking,zhou2025gsm}. Likewise, \textsc{AR-Checker} \citep{hou2025automatic} employs iterative LLM-based rewriting with multi-round verification to automatically generate perturbed math problems across GSM8K, MATH-500, and even general-purpose benchmarks such as MMLU and CommonsenseQA \citep{csqa}. These methods demonstrate that augmenting benchmarks with controlled perturbations can expose failure modes not captured by static test sets \citep{singh2024exposing}. Beyond mathematics, similar frameworks have explored robustness in language understanding and slot-filling \citep{dong2023revisit} by introducing typos, paraphrases, verbosity, or speech-like noise, as well as in commonsense reasoning tasks where definitions are rephrased or critical information is withheld. Collectively, these benchmark augmentation efforts highlight the value of producing harder and more diverse test items. However, they typically optimize against a single model or rely on local perturbations that target narrow error patterns, which can limit transferability and fairness. In contrast, \MODEL evolves benchmarks with explicit ability preservation and multi-model feedback, ensuring that the augmented items are not only more difficult but also more diagnostic and equitable across a diverse model pool.

\stitle{LLM-based Prompt Optimization}
Two related lines of research employ LLMs to optimize prompts. The first focuses on automatic prompt engineering to enhance model performance \citep{liu2025metascale,zhou2022large,yang2023large,pryzant-etal-2023-automatic,deng-etal-2022-rlprompt}. These works share our core idea of leveraging LLMs to automatically reformulate prompts under explicit constraints, ensuring that core conditions are preserved during optimization. Specifically, their goal is to maximize the accuracy of a single target model. In contrast, our objective is to evolve benchmark items that expose failures across a diverse model pool. A second line of work studies jailbreak attacks, which design adversarial prompts that bypass alignment safeguards and elicit harmful content \citep{Perez2022RedTeamLM,Shen2023DAN,Guo2021GBDA,Wen2023PEZ,Wallace2019UAT,Liu2024AutoDAN}. For instance, Prompt Automatic Iterative Refinement (PAIR) \citep{pair_jailbreaking} leverages an LLM-based attacker to reformulate malicious instructions and uses GPT-4 \citep{openai2023gpt4systemcard} as an evaluator to assess the resulting harmful responses. While such approaches target safety violations in a single model, \MODEL uses iterative evolution to update queries in a way that preserves alignment with the original task objective while amplifying difficulty. This distinction highlights our emphasis on benchmark evolution for cross-model robustness and fairness, rather than performance maximization or targeted adversarial failure.

\section{\MODEL}

\MODEL (\Cref{fig:main}) takes as input a benchmark dataset $\mathcal{B} = \{(x_i, y_i)\}_{i=1}^N$ and a pool of $K$ language models $\mathcal{M} = \{M_1, M_2, \dots, M_K\}$ to be evaluated. Each test instance $(x_i, y_i)$ consists of a natural language query $x_i$ and a corresponding reference label $y_i$, and is assumed to assess a well-defined model ability such as mathematical reasoning or safety. \MODEL aims to produce an updated benchmark $\mathcal{B}'$ with improved discriminative power that exposes shared failure patterns, while preserving alignment with the original task objective and ensuring fair evaluation across models.

\subsection{Evaluation Target Extraction}

For each instance $(x_i, y_i)$ in $\mathcal{B}$, we first identify the target ability being evaluated, denoted as $a_i$. This ability description is produced by prompting a language model to summarize the reasoning skill or decision criterion required to solve the input. 
The output $a_i$ is a structured explanation of the current test instance and serves two purposes: it guides the generation of new candidate queries targeting the same competency and provides conditioning context to ensure the evolution process preserves the original evaluation intent.
% and is used to guide both candidate generation and in-context conditioning during benchmark evolution, ensuring that new queries target the same competency and conditioning the generative model accordingly.

\subsection{Candidate Generation and Verification}

Given $(x_i, y_i, a_i)$, we generate a set of $n$ candidate rewrites $\{(x_i^j, y_i^j)\}_{j=1}^{n}$ that preserve the task intent while altering structure or surface form. Each candidate is produced by a conditional language model $G$ with input prompt $\texttt{Prompt}(x_i, y_i, a_i)$ that includes both the original instance and its extracted ability $a_i$. The generator is instructed to preserve the answer validity while introducing controlled variation (e.g., syntactic variation, alternative constraints, or context manipulations) to increase difficulty. To ensure that $y_i^j$ remains the correct answer for $x_i^j$, we verify each candidate using a judgment model $J$, and retain only candidates satisfying $J(x_i^j, y_i^j) = \texttt{Valid}$.

\subsection{Multi-Model Feedback Scoring}

For candidate scoring, we sample a subset $\mathcal{M}_s \subset \mathcal{M}$ of size $m = \lceil \sqrt{K} \rceil$, where $K$ is the total number of available models. Following classical ensemble heuristics \citep{chen2016xgboost,breiman2001random}, the $\sqrt{K}$ rule balances diversity and stability: it yields sufficiently heterogeneous feedback to decorrelate signals while keeping computation tractable. Sampling too few models introduces high-variance, model-idiosyncratic scores, %estimates that are sensitive to individual model behaviors, 
whereas sampling too many %reduces the benefit of diversity and increases the computational cost.
leads to diminished returns on diversity and inflates cost.

Let $\ell(M_k, x)$ denote the loss of model $M_k$ on input $x$, or a task-specific proxy such as inverse log-likelihood or refusal confidence. For candidate $(x_i^j, y_i^j)$, we aggregate feedback across the sampled models by averaging
\begin{equation}
\mathcal{L}(x_i^j) = \frac{1}{m} \sum_{M_k \in \mathcal{M}_s} \ell(M_k, x_i^j) .
\end{equation}

We then select the $k$ candidates with the highest aggregated scores, yielding the updated set
\begin{equation}
\mathcal{X}_i^* = \text{TopK}_{j} \left\{ \mathcal{L}(x_i^j) \right\} .
\end{equation}

\MODEL favors queries that consistently degrade performance across multiple models. By using collective feedback, the selection process avoids overfitting to individual model idiosyncrasies and promotes the discovery of test cases that reflect shared failure modes.

To maintain fairness over the entire benchmark, we enforce near-uniform model sampling. Specifically, we track per-model draw counts throughout all test case updates %on the benchmark. The selection policy is designed such that each model is sampled approximately the same number of times across all iterations. This guarantees balanced exposure and prevents any single model from being overrepresented in the scoring process.
and, at each iteration, preferentially sample under-represented models so that usage converges to parity. This balanced exposure prevents over-representation and keeps the scoring process unbiased.

\subsection{Iterative Refinement with In-context Demonstration}

After selecting the top candidates $\mathcal{X}_i^* = \{(x_i^{(j)}, y_i^{(j)})\}_{j=1}^{k}$, we repurpose them as in-context demonstrations for the next generation round. Each demonstration is constructed by formatting a candidate query and its answer into a standardized template $\texttt{Demo}(x, y)$. The subsequence prompt concatenates the $k$ demonstrations before the original instance and its extracted ability:
\begin{equation*}
\texttt{Prompt}_{\text{next}} = \texttt{Concat}\left(\texttt{Demo}(x_i^{(1)}, y_i^{(1)}), \dots, \texttt{Demo}(x_i^{(k)}, y_i^{(k)}), \texttt{Prompt}(x_i, y_i, a_i)\right) .
\end{equation*}

The test case generator is thus encouraged to produce new queries that preserve the reasoning structure and difficulty profile of the retained candidates while remaining aligned with the target ability. The refinement proceeds for a fixed number of iterations.

\subsection{Final Selection and Benchmark Update}
\label{method:metric}

At the end of the refinement loop, we select the final updated query $x_i^\dagger$ from the last generation stage and assemble the revised benchmark $\mathcal{B}' = \{(x_i^\dagger, y_i) \mid i = 1, \dots, N\}$. We then evaluate the updated benchmark $\mathcal{B}'$ using four quantitative desiderata (with model pool $\mathcal{M}$ of size $K$):

\stitle{Difficulty} Following \cite{li2024autobencher}, a benchmark is considered more difficult if models achieve lower accuracy or higher loss on the updated queries. We define the difficulty of a benchmark $\mathcal{B}'$ with respect to a model pool $\mathcal{M}$ as:
\[
\textsc{Difficulty}(\mathcal{B}', \mathcal{M}) = 1 - \max_{M_k \in \mathcal{M}} \textsc{Acc}(M_k, \mathcal{B}') ,
\]
where $\textsc{Acc}(M_k, \mathcal{B}')$ is the accuracy of model $M_k$ on $\mathcal{B}'$. This metric reflects the remaining headroom for progress by measuring the inverse of the best-performing model's accuracy.

\stitle{Separability} Following \cite{li2024autobencher}, separability measures how well the benchmark spreads model performance. We compute separability as the mean absolute deviation of model accuracies from their mean:
\[
\textsc{Sep}(\mathcal{B}', \mathcal{M}) = \frac{1}{K} \sum_{k=1}^{K} \left| \textsc{Acc}(M_k, \mathcal{B}') - \bar{v} \right|, \quad \text{where } \bar{v} = \frac{1}{K} \sum_{k=1}^{K} \textsc{Acc}(M_k, \mathcal{B}') ,
\]
encouraging settings that avoid near-tied results and sharpen cross-model distinctions.
%This metric ensures the benchmark provides robust comparisons and does not induce identical or near-identical results across models.

\stitle{Fairness}
We assess fairness by measuring how evenly the updated benchmark distributes failure cases across the model pool. For each model $M_k \in \mathcal{M}$, let $c_k$ denote the total number of updated queries $(x_i', y_i')$ on which $M_k$ fails:
\[
c_k = \sum_{i=1}^{|\mathcal{B}'|} \mathbb{I} \left[ \textsc{Fail}(M_k, x_i') \right],
\]
where $\mathbb{I}[\cdot]$ is the indicator function and $\textsc{Fail}(\cdot)$ is a task-specific failure criterion (e.g., incorrect prediction for reasoning tasks or inappropriate generation for safety tasks). Let $\bar{c}$ denote the average number of failures across all models $\bar{c} = \frac{1}{K} \sum_{k=1}^{K} c_k$.
We define fairness as the inverse of the average absolute deviation from this mean, normalized by the total number of benchmark items:
\[
\textsc{Fairness}(\mathcal{B}', \mathcal{M}) = (1 - \frac{ \frac{1}{K} \sum_{k=1}^{K} \left| c_k - \bar{c} \right| }{ |\mathcal{B}'|}) \times 100\%.
\]
This metric encourages updates that reveal shared weaknesses across models, while penalizing benchmarks that disproportionately target only a few specific models.

\stitle{Alignment} We verify alignment via LLM-as-a-judge to ensure that each updated query preserves the core ability and evaluation intent of the original test case. For each test case, we provide the ability description $a_i$, the original question-answer pair, and the updated version. The judge follows a rubric that checks skill equivalence. The alignment score is the proportion of updated items that the judge labels as aligned:
\[
\textsc{Align}(\mathcal{B}') \;=\; \frac{1}{|\mathcal{B}'|}\sum_{i=1}^{|\mathcal{B}'|} \mathbb{I}\!\left[\textsc{Aligned}\!\left(a_i, x_i, y_i, x_i^\dagger, y_i^\dagger\right)\right].
\]

\begin{algorithm}[t]
\caption{\MODEL}
\label{alg:arenabencher}
\begin{algorithmic}[1]
\Require Benchmark $\mathcal{B} = \{(x_i, y_i)\}_{i=1}^N$, model pool $\mathcal{M} = \{M_1, \dots, M_K\}$, number of candidates per query $n$, refinement rounds $R$
\Ensure Updated benchmark $\mathcal{B}' = \{(x_i^\dagger, y_i^\dagger)\}_{i=1}^N$
\For{each $(x_i, y_i)$ in $\mathcal{B}$}
    \State Extract ability description $a_i \gets \textsc{AbilityDesc}(x_i, y_i)$
    \State Initialize prompt $p \gets (x_i, y_i, a_i)$
    \For{$r = 1$ to $R$}
        \State Generate $n$ candidates $\{(x_i^j, y_i^j)\}_{j=1}^{n}$ using generator $G$ with prompt $p$
        \State Filter invalid $(x_i^j, y_i^j)$ using verifier $J$: keep only if $J(x_i^j, y_i^j) = \texttt{Valid}$
        \State Sample model subset $\mathcal{M}_s \subset \mathcal{M}$, size $m = \lceil \sqrt{K} \rceil$
        \For{each valid candidate $(x_i^j, y_i^j)$}
            \State Compute loss $\mathcal{L}(x_i^j, y_i^j) = \frac{1}{m} \sum_{M_k \in \mathcal{M}_s} \ell(M_k, x_i^j, y_i^j)$
        \EndFor
        \State Select top-$k$ candidates $\mathcal{X}_i^* \gets \text{TopK}_{j} \{ \mathcal{L}(x_i^j, y_i^j) \}$
        \State Update prompt $p \gets \textsc{Demo}(\mathcal{X}_i^*) \cup (x_i, y_i, a_i)$
    \EndFor
    \State Select final pair $(x_i^\dagger, y_i^\dagger) \gets \arg\max_{(x, y) \in \mathcal{X}_i^*} \mathcal{L}(x, y)$
\EndFor
\State \Return $\mathcal{B}' = \{(x_i^\dagger, y_i^\dagger)\}_{i=1}^N$
\end{algorithmic}
\end{algorithm}

\section{Experiments and Evaluation}

\begin{table}[t]
\centering
\caption{Performance of \MODEL on three representative tasks: GSM8K (math), Harmful Behaviors (safety), and CSQA (CommonsenseQA, reasoning). $m$ denotes the number of models sampled to gather feedback for each query update. $Acc (\uparrow)$ indicates accuracy, $ASR (\downarrow)$ indicates attack success rate, and $\Delta$ indicates the change after benchmark update. Model names with the ``-I'' suffix (e.g., Llama-3.2-3B-I) indicate instruction-tuned variants.}
\vspace{0.8em}
\label{tab:main}
\renewcommand{\arraystretch}{1.2}
\resizebox{\linewidth}{!}{%
\begin{tabular}{lclllllllll}
\toprule
\multirow{2}{*}{Model Pool} & \multirow{2}{*}{\#$m$} 
& \multicolumn{3}{c}{GSM8K} 
& \multicolumn{3}{c}{Harmful Behaviors} 
& \multicolumn{3}{c}{CSQA} \\
\cmidrule(l){3-5} \cmidrule(l){6-8} \cmidrule(l){9-11}
 &  & $Acc_{ori}$ & $Acc_{up}$ & $\Delta Acc$ & $ASR_{ori}$ & $ASR_{up}$ & $\Delta ASR$ & $Acc_{ori}$ & $Acc_{up}$ & $\Delta Acc$ \\
\midrule
\rowcolor{gray!10}
\multirow{2}{*}{\cellcolor{white}Llama-3.2-1B} & 3 & \multirow{2}{*}{44.4} & 12.9 & $\color{teal}{\downarrow}$ 31.5 & \multirow{2}{*}{67.8} & 76.4 & $\color{red}{\uparrow}$ 8.6 & \multirow{2}{*}{42.1} & 22.3 & $\color{teal}{\downarrow}$ 19.8 \\
 & 1 & & 22.1 & $\color{teal}{\downarrow}$ 22.3 & & 73.2 & $\color{red}{\uparrow}$ 5.4 &   & 26.7 & $\color{teal}{\downarrow}$ 15.4  \\
\rowcolor{gray!10}
\multirow{2}{*}{\cellcolor{white}Llama-3.2-3B} & 3 & \multirow{2}{*}{74.1} & 26.4 & $\color{teal}{\downarrow}$ 47.7 & \multirow{2}{*}{54.6} & 68.2 & $\color{red}{\uparrow}$ 13.6 & \multirow{2}{*}{60.6} & 32.0 & $\color{teal}{\downarrow}$ 28.6 \\
 & 1 &  & 41.3 & $\color{teal}{\downarrow}$ 32.8 &  & 62.6 & $\color{red}{\uparrow}$ 8.0 &  & 48.1 & $\color{teal}{\downarrow}$ 12.5 \\
\rowcolor{gray!10}
\multirow{2}{*}{\cellcolor{white}Llama-3.2-3B-I} & 3 & \multirow{2}{*}{78.3} & 38.1 & $\color{teal}{\downarrow}$ 40.2 & \multirow{2}{*}{29.4} & 43.4 & $\color{red}{\uparrow}$ 14.0 & \multirow{2}{*}{65.1} & 53.0 & $\color{teal}{\downarrow}$ 12.1 \\
 & 1 &  & 43.7 & $\color{teal}{\downarrow}$ 34.6 &  & 30.8 & $\color{red}{\uparrow}$ 1.4 &  & 56.7 & $\color{teal}{\downarrow}$ 8.4 \\
\rowcolor{gray!10}
\multirow{2}{*}{\cellcolor{white}Qwen3-4B} & 3 & \multirow{2}{*}{87.8} & 52.1 & $\color{teal}{\downarrow}$ 35.7 & \multirow{2}{*}{5.2} & 24.2 & $\color{red}{\uparrow}$ 19.0 & \multirow{2}{*}{53.0} & 26.7 & $\color{teal}{\downarrow}$ 26.3 \\
 & 1 &  & 63.7 & $\color{teal}{\downarrow}$ 24.1 & & 13.4 & $\color{red}{\uparrow}$ 8.2 & & 32.0  & $\color{teal}{\downarrow}$ 21.0 \\
\rowcolor{gray!10}
\multirow{2}{*}{\cellcolor{white}Qwen3-4B-I} & 3 & \multirow{2}{*}{90.1} & 58.6 & $\color{teal}{\downarrow}$ 31.5 & \multirow{2}{*}{33.4} & 44.6 & $\color{red}{\uparrow}$ 11.2 & \multirow{2}{*}{68.6} & 33.9 & $\color{teal}{\downarrow}$ 34.7 \\
 & 1 &  & 63.7 & $\color{teal}{\downarrow}$ 26.4 &  & 42.2 & $\color{red}{\uparrow}$ 8.8 &  & 42.4 & $\color{teal}{\downarrow}$ 26.2 \\
\rowcolor{gray!10}
\multirow{2}{*}{\cellcolor{white}Mistral-7B-I} & 3 & \multirow{2}{*}{52.2} & 39.4 & $\color{teal}{\downarrow}$ 12.8 & \multirow{2}{*}{22.6} & 46.2 & $\color{red}{\uparrow}$ 23.6 & \multirow{2}{*}{48.1} & 28.4 & $\color{teal}{\downarrow}$ 19.7 \\
 & 1 &  & 46.2 & $\color{teal}{\downarrow}$ 6.0 &  & 24.8 & $\color{red}{\uparrow}$ 2.2 &  & 32.0 & $\color{teal}{\downarrow}$ 16.1 \\
\bottomrule
\end{tabular}
}
\end{table}

\subsection{Experimental Settings}

\stitle{Models and Benchmarks}
We evaluate \MODEL on a diverse pool of open-source language models to ensure that the observed effects are not confined to a single architecture or training pipeline. Concretely, we include three representative families of models: \textbf{LLaMA3} \citep{grattafiori2024llama}, \textbf{Qwen3} \citep{yang2024qwen2technicalreport}, and \textbf{Mistral} \citep{jiang2024mistral}. For each family, we select both the base and instruction-tuned variants, covering parameter scales from 1B to 4B. The complete model list is shown in \Cref{tab:main}. To comprehensively examine adaptability and generalizability, we apply \MODEL to three domains. For safety, we use the AdvBench Harmful Behaviors dataset \citep{zou2023universal}, which consists of prompts designed to elicit unsafe or malicious outputs. The goal is to evaluate whether \MODEL can update safety benchmarks to continue surfacing vulnerabilities that models fail to reject, even as they become more resistant to obvious unsafe requests. For mathematical reasoning, we adopt GSM8K \citep{gsm8k}, a widely used benchmark for multi-step arithmetic that requires decomposition and intermediate reasoning. This setting tests whether \MODEL can generate updated queries that remain solvable by the same ground-truth answers while significantly increasing difficulty and exposing reasoning failures. For commonsense reasoning, we use CommonsenseQA \citep{csqa}, which evaluates models on everyday inferential reasoning beyond surface-level linguistic cues. This benchmark enables us to assess whether \MODEL can produce variations that continue to test the same commonsense skills while presenting more nuanced or less frequent contexts. For each benchmark, we evaluate all models on both the original dataset and the updated dataset evolved by \MODEL.

\stitle{Metrics}
Our evaluation is based on the metrics introduced in \Cref{method:metric}. In the safety domain, we measure the \emph{attack success rate} (ASR), defined as the proportion of adversarial prompts that induce unsafe or harmful outputs. A lower ASR indicates stronger refusal ability and greater robustness. In other domains, we report model \emph{accuracy} on the updated benchmarks. We evaluate benchmark quality using four metrics: \emph{fairness}, \emph{separability}, \emph{alignment}, and \emph{difficulty}. 
% Accuracy reflects task success, consistency quantifies the stability of performance across different benchmark variants, and fairness measures whether performance degradation is distributed evenly across models rather than concentrated on a subset of systems.
Fairness measures whether performance degradation is evenly distributed across models rather than concentrated on a few; separability captures how well the updated benchmark distinguishes between models of different capabilities; alignment reflects whether updated queries preserve the original intent or skill coverage; and difficulty quantifies the average performance across models.

\stitle{Hyperparameters}
In each domain, we initialize with an existing benchmark $\mathcal{B}$ and iteratively propose candidate rewrites guided by feedback from a sampled subset of target models. For each iteration, we randomly sample $3$ models from the full model pool $\mathcal{M}$ of size $6$, and use their responses to guide benchmark updates. We run $R = 3$ adaptive iterations, each proposing $n = 5$ candidate generations for a batch of original examples and maintain the top three samples as the in-context demonstration for the next iteration. To ensure fairness across models, \MODEL tracks model sampling frequency and enforces uniform coverage over the full benchmark construction process. We use GPT-4o-2024-08-06 for test objective extraction, test case generation, and as the verifier.

\begin{table}[t]
\centering
\caption{Comparison of benchmark quality metrics across three tasks. We report \emph{Fairness}, \emph{Separability} (\emph{Sep}), \emph{Alignment} (\emph{Align}), and \emph{Difficulty} (\emph{Diff}).
``Ori.'' refers to the original form of the evaluated benchmark. $\MODEL_1$ and $\MODEL_3$ denote \MODEL variants using $m{=}1$ and $m{=}3$ feedback models respectively.}
\vspace{0.8em}
\label{tab:metrics}
\renewcommand{\arraystretch}{1.2}
\resizebox{\linewidth}{!}{%
\begin{tabular}{lcccccccccccc}
\toprule
\multirow{2}{*}{Benchmark} 
& \multicolumn{4}{c}{GSM8K} 
& \multicolumn{4}{c}{Harmful Behaviors} 
& \multicolumn{4}{c}{CSQA} \\
\cmidrule(lr){2-5} \cmidrule(lr){6-9} \cmidrule(lr){10-13}
 & \emph{Fairness} & \emph{Sep} & \emph{Align} & \emph{Diff} 
 & \emph{Fairness} & \emph{Sep} & \emph{Align} & \emph{Diff} 
 & \emph{Fairness} & \emph{Sep} & \emph{Align} & \emph{Diff} \\
\midrule
Ori. & 84.8 & 15.2 & -- & 9.9 & 82.9 & 17.1 & -- & 5.2 & 91.4 & 8.5 & -- & 31.4 \\
$\MODEL_1$ & 88.7 & 11.3 & 94.1 & 36.3 & 81.8 & 18.2 & 92.4 & 13.4 & 90.6 & 9.4 & 93.7 & 43.3 \\
$\MODEL_3$ & 87.8 & 12.2 & 91.3 & 41.4 & 85.47 & 14.5 & 90.6 & 24.2 & 92.8 & 7.2 & 91.4 & 47.0 \\
\bottomrule
\end{tabular}
}
\end{table}

\subsection{Main Results}
\stitle{Overall Performance of \MODEL}
\Cref{tab:main} presents the model performance on the original and updated benchmarks across three distinct domains. We report accuracy ($Acc$) for GSM8K and CommonsenseQA, and attack success rate ($ASR$) for the Harmful Behaviors Dataset. \MODEL consistently increases the difficulty of all benchmarks, as evidenced by the substantial drops in accuracy and rises in ASR. For instance, the LLaMA-3.2-3B model experiences a $47.7\%$ drop in GSM8K accuracy and a $13.6\%$ increase in ASR (according to the $m=3$ setting, which serves as the default configuration of \MODEL), indicating that \MODEL-generated updates effectively expose reasoning gaps and safety vulnerabilities. This trend holds across all model families and domains. Qwen3-4B, for example, shows a $35.7\%$ drop in GSM8K accuracy and a $19.0\%$ ASR increase. Even models that initially exhibit high robustness, such as Mistral-7B-I and Qwen3-4B-I, still suffer notable degradations after benchmark updates.

Notably, instruction-tuned models (e.g., LLaMA-3.2-3B-I and Qwen3-4B-I) are generally more robust than their base counterparts, but nonetheless exhibit substantial performance drops. This suggests that safety-aligned or instruction-tuned models retain exploitable weaknesses that can be surfaced by \MODEL's targeted updates. Overall, these results demonstrate that \MODEL can effectively evolve existing benchmarks to produce updated datasets that are more difficult, safety-sensitive, and diagnostic of model limitations.

\stitle{Improved Benchmark Quality}
To better understand the properties of the updated benchmarks generated by \MODEL, we evaluate them using four complementary metrics: \emph{Fairness}, \emph{Separability}, \emph{Alignment}, and \emph{Difficulty}. \MODEL substantially improves benchmark quality across all domains, as shown in \Cref{tab:metrics}. \MODEL substantially improves benchmark quality across all three domains. The \emph{difficulty} of the updated benchmarks increases markedly, indicating that the generated queries are meaningfully harder for models and reveal more failure cases. At the same time, \emph{alignment} remains consistently high, showing that the updates preserve the original task intent and focus, such as math reasoning or commonsense inference. \emph{Fairness} also improves or stays stable, suggesting that performance degradation is more evenly distributed across models, rather than disproportionately affecting a few. While \emph{separability} experiences slight variation, this is expected as model performance begins to compress under increased difficulty. Nonetheless, the updated benchmarks maintain sufficient variance to differentiate model capabilities. These results demonstrate that \MODEL produces benchmark updates that are more challenging, semantically faithful, fair across systems, and still diagnostic of model differences under stress.

% \subsection{Analysis}

\stitle{Effect of Multi-model Feedback}
We compare the performance of \MODEL under two configurations, where feedback is collected from either a single model ($m{=}1$) or from $\lceil \sqrt{K} \rceil$ models ($m{=}3$) during each test case update. As shown in \Cref{tab:main}, using multiple feedback models consistently results in greater performance degradation across all tasks. For example, across most model families, the drop in accuracy on GSM8K and CSQA is larger under $m{=}3$ than under $m{=}1$. Similarly, the attack success rate increases more under $m{=}3$ in the safety domain. These trends suggest that aggregating signals from multiple models leads to more effective query updates that are harder for all models in the pool.
We further examine benchmark quality metrics in \Cref{tab:metrics}. The $m{=}3$ configuration produces benchmarks with higher difficulty across all domains. Fairness and alignment remain strong and are comparable to the $m{=}1$ setting, indicating that performance degradation remains well distributed and semantically consistent even when updates are guided by multiple models. Separability varies slightly between the two settings, but remains within a comparable range, suggesting that the ability to distinguish model capabilities is preserved.

\begin{figure}[t]
   \centering 
   \includegraphics[width=0.95\textwidth]{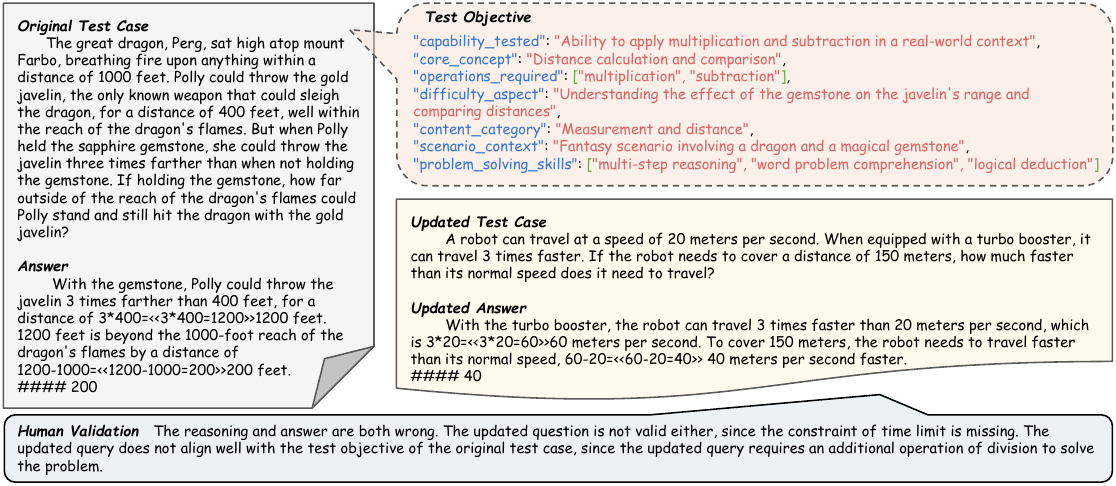}
   \caption{Case study of \MODEL-generated test case update. While the objective extraction succeeds, the updated test case generated by \MODEL fails for two key reasons. First, the updated question is not well-formed, as it omits necessary information, making it unsolvable. Second, although the updated query retains a similar surface-level objective, it introduces additional complexity by requiring a new mathematical operation (division), thus deviating from the original reasoning structure and increasing cognitive load.}
    \label{fig:case}
\end{figure}

\stitle{Human Annotation}
To further validate the quality of benchmark updates beyond automatic metrics, we conduct human evaluation on $100$ randomly sampled updated test cases from GSM8K. The samples are annotated independently by three expert annotators with sufficient expertise in mathematics. Each sample is evaluated along two axes: \emph{alignment}, which measures whether the updated query preserves the original test objective, and \emph{correctness}, which assesses whether the question and answer pair is valid and solvable. Among the $100$ annotated samples, $95$ are judged as aligned with the original intent, and $96$ are considered correct in terms of question formulation and answer validity. These results confirm that \MODEL not only increases benchmark difficulty, but also preserves semantic fidelity and ensures correctness in the majority of cases.

\stitle{Case Study}
Although the combination of test objective extraction and the verifier aims to ensure the correctness and alignment of each benchmark update, failure cases can still arise. \Cref{fig:case} presents a failure case that highlights the challenges in maintaining semantic fidelity during benchmark evolution. The original test case involves a fantasy scenario requiring multi-step reasoning over distances, centered around multiplication and subtraction. From this, \MODEL extracts a structured test objective capturing the intended reasoning capability, core concept, and scenario context. The extracted objective is accurate and reflects the underlying skill being tested. However, the updated test case generated by \MODEL fails in two important ways. First, the revised query is not valid as a standalone question. It omits the essential time constraint needed to perform a valid speed–distance comparison, rendering the question underspecified and unsolvable. Second, although the surface structure of the test objective appears preserved, the updated version introduces an additional operation of division, increasing the overall reasoning complexity. As a result, the updated query no longer faithfully tests the same skill profile as the original.

\section{Conclusion and Future Work}
In this paper, we present \MODEL, a framework for automatic benchmark evolution via multi-model competitive evaluation. Given an existing benchmark and a diverse pool of target language models, \MODEL infers the core ability of each test case, generates aligned variants with a language model, verifies answer correctness and intent with an independent judge, and selects candidates that consistently degrade performance across multiple models. The framework maintains an in-context memory of challenging examples to guide subsequent updates. Experiments on GSM8K, CommonsenseQA, and a safety dataset show that \MODEL increases difficulty while preserving alignment and fairness, and largely maintains separability. \MODEL is a first step toward continuously evolving and contamination-resilient evaluation that uses multi-model signals to generate and evolve test cases. Future work will broaden the scope to multimodal settings and strengthen validity checks with structure-aware constraints and ensembles of calibrated judges.

% \input{sections/7_limitations}

% \input{sections/8_ethics_statements}

% \subsubsection*{Author Contributions}
% If you'd like to, you may include  a section for author contributions as is done
% in many journals. This is optional and at the discretion of the authors.

% \subsubsection*{Acknowledgments}
% Use unnumbered third level headings for the acknowledgments. All
% acknowledgments, including those to funding agencies, go at the end of the paper.

\bibliography{iclr2026_conference}

\begin{thebibliography}{61}
\providecommand{\natexlab}[1]{#1}
\providecommand{\url}[1]{\texttt{#1}}
\expandafter\ifx\csname urlstyle\endcsname\relax
  \providecommand{\doi}[1]{doi: #1}\else
  \providecommand{\doi}{doi: \begingroup \urlstyle{rm}\Url}\fi

\bibitem[Abedin et~al.(2025)Abedin, Qamar, Flek, and Karimi]{abedin2025arithmattack}
Zain~Ul Abedin, Shahzeb Qamar, Lucie Flek, and Akbar Karimi.
\newblock Arithmattack: Evaluating robustness of llms to noisy context in math problem solving.
\newblock \emph{arXiv preprint arXiv:2501.08203}, 2025.

\bibitem[Bai et~al.(2022)Bai, Kadavath, Kundu, Askell, Kernion, Jones, Chen, Goldie, Mirhoseini, McKinnon, et~al.]{bai2022constitutional}
Yuntao Bai, Saurav Kadavath, Sandipan Kundu, Amanda Askell, Jackson Kernion, Andy Jones, Anna Chen, Anna Goldie, Azalia Mirhoseini, Cameron McKinnon, et~al.
\newblock Constitutional ai: Harmlessness from ai feedback.
\newblock \emph{arXiv preprint arXiv:2212.08073}, 2022.

\bibitem[Balloccu et~al.(2024)Balloccu, Schmidtov{\'a}, Lango, and Dusek]{balloccu-etal-2024-leak}
Simone Balloccu, Patr{\'i}cia Schmidtov{\'a}, Mateusz Lango, and Ondrej Dusek.
\newblock Leak, cheat, repeat: Data contamination and evaluation malpractices in closed-source {LLM}s.
\newblock In Yvette Graham and Matthew Purver (eds.), \emph{Proceedings of the 18th Conference of the European Chapter of the Association for Computational Linguistics (Volume 1: Long Papers)}, pp.\  67--93, St. Julian{'}s, Malta, March 2024. Association for Computational Linguistics.
\newblock \doi{10.18653/v1/2024.eacl-long.5}.
\newblock URL \url{https://aclanthology.org/2024.eacl-long.5/}.

\bibitem[Breiman(2001)]{breiman2001random}
Leo Breiman.
\newblock Random forests.
\newblock \emph{Machine learning}, 45\penalty0 (1):\penalty0 5--32, 2001.

\bibitem[Chao et~al.(2025)Chao, Robey, Dobriban, Hassani, Pappas, and Wong]{pair_jailbreaking}
Patrick Chao, Alexander Robey, Edgar Dobriban, Hamed Hassani, George~J Pappas, and Eric Wong.
\newblock Jailbreaking black box large language models in twenty queries.
\newblock In \emph{2025 IEEE Conference on Secure and Trustworthy Machine Learning (SaTML)}, pp.\  23--42. IEEE, 2025.

\bibitem[Chen et~al.(2025)Chen, Chen, Li, Jiang, Wan, He, Ran, Gu, Li, Xie, et~al.]{benchmark_contamination_advancements}
Simin Chen, Yiming Chen, Zexin Li, Yifan Jiang, Zhongwei Wan, Yixin He, Dezhi Ran, Tianle Gu, Haizhou Li, Tao Xie, et~al.
\newblock Recent advances in large langauge model benchmarks against data contamination: From static to dynamic evaluation.
\newblock \emph{arXiv preprint arXiv:2502.17521}, 2025.

\bibitem[Chen \& Guestrin(2016)Chen and Guestrin]{chen2016xgboost}
Tianqi Chen and Carlos Guestrin.
\newblock Xgboost: A scalable tree boosting system.
\newblock In \emph{Proceedings of the 22nd acm sigkdd international conference on knowledge discovery and data mining}, pp.\  785--794, 2016.

\bibitem[Choi et~al.(2025)Choi, Khanov, Wei, and Li]{choicontaminated}
Hyeong~Kyu Choi, Maxim Khanov, Hongxin Wei, and Yixuan Li.
\newblock How contaminated is your benchmark? measuring dataset leakage in large language models with kernel divergence.
\newblock In \emph{Forty-second International Conference on Machine Learning}, 2025.

\bibitem[Cobbe et~al.(2021)Cobbe, Kosaraju, Bavarian, Chen, Jun, Kaiser, Plappert, Tworek, Hilton, Nakano, et~al.]{gsm8k}
Karl Cobbe, Vineet Kosaraju, Mohammad Bavarian, Mark Chen, Heewoo Jun, Lukasz Kaiser, Matthias Plappert, Jerry Tworek, Jacob Hilton, Reiichiro Nakano, et~al.
\newblock Training verifiers to solve math word problems.
\newblock \emph{arXiv preprint arXiv:2110.14168}, 2021.

\bibitem[Dekoninck et~al.(2024)Dekoninck, M{\"u}ller, Baader, Fischer, and Vechev]{dekoninck2024evading}
Jasper Dekoninck, Mark~Niklas M{\"u}ller, Maximilian Baader, Marc Fischer, and Martin Vechev.
\newblock Evading data contamination detection for language models is (too) easy.
\newblock \emph{arXiv preprint arXiv:2402.02823}, 2024.

\bibitem[Deng et~al.(2022)Deng, Wang, Hsieh, Wang, Guo, Shu, Song, Xing, and Hu]{deng-etal-2022-rlprompt}
Mingkai Deng, Jianyu Wang, Cheng-Ping Hsieh, Yihan Wang, Han Guo, Tianmin Shu, Meng Song, Eric Xing, and Zhiting Hu.
\newblock {RLP}rompt: Optimizing discrete text prompts with reinforcement learning.
\newblock In Yoav Goldberg, Zornitsa Kozareva, and Yue Zhang (eds.), \emph{Proceedings of the 2022 Conference on Empirical Methods in Natural Language Processing}, pp.\  3369--3391, Abu Dhabi, United Arab Emirates, December 2022. Association for Computational Linguistics.
\newblock \doi{10.18653/v1/2022.emnlp-main.222}.
\newblock URL \url{https://aclanthology.org/2022.emnlp-main.222/}.

\bibitem[Dong et~al.(2023)Dong, Zhao, Hui, Guo, Wang, Feng, Qiu, Gongque, He, Wang, et~al.]{dong2023revisit}
Guanting Dong, Jinxu Zhao, Tingfeng Hui, Daichi Guo, Wenlong Wang, Boqi Feng, Yueyan Qiu, Zhuoma Gongque, Keqing He, Zechen Wang, et~al.
\newblock Revisit input perturbation problems for llms: A unified robustness evaluation framework for noisy slot filling task.
\newblock In \emph{CCF International Conference on Natural Language Processing and Chinese Computing}, pp.\  682--694. Springer, 2023.

\bibitem[Dong et~al.(2024)Dong, Jiang, Liu, Jin, Gu, Yang, and Li]{dong-etal-2024-generalization}
Yihong Dong, Xue Jiang, Huanyu Liu, Zhi Jin, Bin Gu, Mengfei Yang, and Ge~Li.
\newblock Generalization or memorization: Data contamination and trustworthy evaluation for large language models.
\newblock In Lun-Wei Ku, Andre Martins, and Vivek Srikumar (eds.), \emph{Findings of the Association for Computational Linguistics: ACL 2024}, pp.\  12039--12050, Bangkok, Thailand, August 2024. Association for Computational Linguistics.
\newblock \doi{10.18653/v1/2024.findings-acl.716}.
\newblock URL \url{https://aclanthology.org/2024.findings-acl.716/}.

\bibitem[Fan et~al.(2023)Fan, Hua, Li, Ling, and Zhang]{fan2023nphardeval}
Lizhou Fan, Wenyue Hua, Lingyao Li, Haoyang Ling, and Yongfeng Zhang.
\newblock Nphardeval: Dynamic benchmark on reasoning ability of large language models via complexity classes.
\newblock \emph{arXiv preprint arXiv:2312.14890}, 2023.

\bibitem[Grattafiori et~al.(2024)Grattafiori, Dubey, Jauhri, Pandey, Kadian, Al-Dahle, Letman, Mathur, Schelten, Vaughan, et~al.]{grattafiori2024llama}
Aaron Grattafiori, Abhimanyu Dubey, Abhinav Jauhri, Abhinav Pandey, Abhishek Kadian, Ahmad Al-Dahle, Aiesha Letman, Akhil Mathur, Alan Schelten, Alex Vaughan, et~al.
\newblock The llama 3 herd of models.
\newblock \emph{arXiv preprint arXiv:2407.21783}, 2024.

\bibitem[Gu et~al.(2024)Gu, Jiang, Shi, Tan, Zhai, Xu, Li, Shen, Ma, Liu, et~al.]{gu2024survey}
Jiawei Gu, Xuhui Jiang, Zhichao Shi, Hexiang Tan, Xuehao Zhai, Chengjin Xu, Wei Li, Yinghan Shen, Shengjie Ma, Honghao Liu, et~al.
\newblock A survey on llm-as-a-judge.
\newblock \emph{arXiv preprint arXiv:2411.15594}, 2024.

\bibitem[Guo et~al.(2021)Guo, Sablayrolles, J'egou, and Kiela]{Guo2021GBDA}
Chuan Guo, Alexandre Sablayrolles, Herv'e J'egou, and Douwe Kiela.
\newblock Gradient-based adversarial attacks against text transformers.
\newblock In \emph{Conference on Empirical Methods in Natural Language Processing}, 2021.
\newblock URL \url{https://api.semanticscholar.org/CorpusID:233423658}.

\bibitem[Hendrycks et~al.(2020)Hendrycks, Burns, Basart, Zou, Mazeika, Song, and Steinhardt]{mmlu}
Dan Hendrycks, Collin Burns, Steven Basart, Andy Zou, Mantas Mazeika, Dawn Song, and Jacob Steinhardt.
\newblock Measuring massive multitask language understanding.
\newblock \emph{arXiv preprint arXiv:2009.03300}, 2020.

\bibitem[Hendrycks et~al.(2021)Hendrycks, Burns, Kadavath, Arora, Basart, Tang, Song, and Steinhardt]{math-500}
Dan Hendrycks, Collin Burns, Saurav Kadavath, Akul Arora, Steven Basart, Eric Tang, Dawn Song, and Jacob Steinhardt.
\newblock Measuring mathematical problem solving with the math dataset.
\newblock \emph{arXiv preprint arXiv:2103.03874}, 2021.

\bibitem[Hong et~al.(2024)Hong, Majumder, Ghosal, Aditya, Mihalcea, and Poria]{hong2024evaluating}
Pengfei Hong, Navonil Majumder, Deepanway Ghosal, Somak Aditya, Rada Mihalcea, and Soujanya Poria.
\newblock Evaluating llms' mathematical and coding competency through ontology-guided interventions.
\newblock \emph{arXiv preprint arXiv:2401.09395}, 2024.

\bibitem[Hou et~al.(2025)Hou, Xiao, Yu, Jiang, Wei, Huang, Chen, and Chen]{hou2025automatic}
Yutao Hou, Zeguan Xiao, Fei Yu, Yihan Jiang, Xuetao Wei, Hailiang Huang, Yun Chen, and Guanhua Chen.
\newblock Automatic robustness stress testing of llms as mathematical problem solvers.
\newblock \emph{arXiv preprint arXiv:2506.05038}, 2025.

\bibitem[Huang et~al.(2025)Huang, Guo, Li, Ji, Ge, Li, Guo, Cai, Yuan, Wang, et~al.]{math_perturb}
Kaixuan Huang, Jiacheng Guo, Zihao Li, Xiang Ji, Jiawei Ge, Wenzhe Li, Yingqing Guo, Tianle Cai, Hui Yuan, Runzhe Wang, et~al.
\newblock Math-perturb: Benchmarking llms' math reasoning abilities against hard perturbations.
\newblock \emph{arXiv preprint arXiv:2502.06453}, 2025.

\bibitem[{Hugging Face}(2023)]{open-llm-leaderboard}
{Hugging Face}.
\newblock Open {LLM} leaderboard.
\newblock \url{https://huggingface.co/spaces/open-llm-leaderboard/open_llm_leaderboard}, 2023.
\newblock Accessed: [Your access date].

\bibitem[Jain et~al.(2024)Jain, Han, Gu, Li, Yan, Zhang, Wang, Solar-Lezama, Sen, and Stoica]{jain2024livecodebench}
Naman Jain, King Han, Alex Gu, Wen-Ding Li, Fanjia Yan, Tianjun Zhang, Sida Wang, Armando Solar-Lezama, Koushik Sen, and Ion Stoica.
\newblock Livecodebench: Holistic and contamination free evaluation of large language models for code.
\newblock \emph{arXiv preprint arXiv:2403.07974}, 2024.

\bibitem[Jiang et~al.(2024{\natexlab{a}})Jiang, Sablayrolles, Mensch, Bamford, Chaplot, Casas, Bressand, Lengyel, Lample, Saulnier, et~al.]{jiang2024mistral}
AQ~Jiang, A~Sablayrolles, A~Mensch, C~Bamford, DS~Chaplot, Ddl Casas, F~Bressand, G~Lengyel, G~Lample, L~Saulnier, et~al.
\newblock Mistral 7b. arxiv 2023.
\newblock \emph{arXiv preprint arXiv:2310.06825}, 2024{\natexlab{a}}.

\bibitem[Jiang et~al.(2024{\natexlab{b}})Jiang, Liu, Zhong, Schaeffer, Ouyang, Han, and Koyejo]{jiang2024investigatingdatacontaminationpretraining}
Minhao Jiang, Ken~Ziyu Liu, Ming Zhong, Rylan Schaeffer, Siru Ouyang, Jiawei Han, and Sanmi Koyejo.
\newblock Investigating data contamination for pre-training language models, 2024{\natexlab{b}}.
\newblock URL \url{https://arxiv.org/abs/2401.06059}.

\bibitem[Lee et~al.(2023)Lee, Phatale, Mansoor, Mesnard, Ferret, Lu, Bishop, Hall, Carbune, Rastogi, et~al.]{lee2023rlaif}
Harrison Lee, Samrat Phatale, Hassan Mansoor, Thomas Mesnard, Johan Ferret, Kellie Lu, Colton Bishop, Ethan Hall, Victor Carbune, Abhinav Rastogi, et~al.
\newblock Rlaif vs. rlhf: Scaling reinforcement learning from human feedback with ai feedback.
\newblock \emph{arXiv preprint arXiv:2309.00267}, 2023.

\bibitem[Li et~al.(2025)Li, Kaiyom, Liu, Mai, Liang, and Hashimoto]{li2024autobencher}
Xiang~Lisa Li, Farzaan Kaiyom, Evan~Zheran Liu, Yifan Mai, Percy Liang, and Tatsunori Hashimoto.
\newblock Autobencher: Towards declarative benchmark construction.
\newblock In \emph{The Thirteenth International Conference on Learning Representations}, 2025.
\newblock URL \url{https://openreview.net/forum?id=ymt4crbbXh}.

\bibitem[Li et~al.(2024)Li, Ma, Ballesteros, Benajiba, and Horwood]{li2024active}
Yang Li, Jie Ma, Miguel Ballesteros, Yassine Benajiba, and Graham Horwood.
\newblock Active evaluation acquisition for efficient llm benchmarking.
\newblock \emph{arXiv preprint arXiv:2410.05952}, 2024.

\bibitem[Liang et~al.(2022)Liang, Bommasani, Lee, Tsipras, Soylu, Yasunaga, Zhang, Narayanan, Wu, Kumar, et~al.]{helm}
Percy Liang, Rishi Bommasani, Tony Lee, Dimitris Tsipras, Dilara Soylu, Michihiro Yasunaga, Yian Zhang, Deepak Narayanan, Yuhuai Wu, Ananya Kumar, et~al.
\newblock Holistic evaluation of language models.
\newblock \emph{arXiv preprint arXiv:2211.09110}, 2022.

\bibitem[Liang et~al.(2025)Liang, Yu, Zhang, Ye, and Hu]{liang2025largelanguagemodelcheat}
Zi~Liang, Liantong Yu, Shiyu Zhang, Qingqing Ye, and Haibo Hu.
\newblock How much do large language model cheat on evaluation? benchmarking overestimation under the one-time-pad-based framework, 2025.
\newblock URL \url{https://arxiv.org/abs/2507.19219}.

\bibitem[Liu et~al.(2025)Liu, Zhou, Xu, Huang, Wang, Zhang, Poon, and Chen]{liu2025metascale}
Qin Liu, Wenxuan Zhou, Nan Xu, James~Y Huang, Fei Wang, Sheng Zhang, Hoifung Poon, and Muhao Chen.
\newblock Metascale: Test-time scaling with evolving meta-thoughts.
\newblock \emph{arXiv preprint arXiv:2503.13447}, 2025.

\bibitem[Liu et~al.(2023)Liu, Xu, Chen, and Xiao]{Liu2024AutoDAN}
Xiaogeng Liu, Nan Xu, Muhao Chen, and Chaowei Xiao.
\newblock Autodan: Generating stealthy jailbreak prompts on aligned large language models.
\newblock \emph{ArXiv}, abs/2310.04451, 2023.
\newblock URL \url{https://api.semanticscholar.org/CorpusID:263831566}.

\bibitem[Mirzadeh et~al.(2024)Mirzadeh, Alizadeh, Shahrokhi, Tuzel, Bengio, and Farajtabar]{gsm_symbolic}
Iman Mirzadeh, Keivan Alizadeh, Hooman Shahrokhi, Oncel Tuzel, Samy Bengio, and Mehrdad Farajtabar.
\newblock Gsm-symbolic: Understanding the limitations of mathematical reasoning in large language models.
\newblock \emph{arXiv preprint arXiv:2410.05229}, 2024.

\bibitem[Mo et~al.(2025)Mo, Liu, Wen, Jung, Askari, Zhou, Zhao, and Chen]{mo2025redcoderautomatedmultiturnred}
Wenjie~Jacky Mo, Qin Liu, Xiaofei Wen, Dongwon Jung, Hadi Askari, Wenxuan Zhou, Zhe Zhao, and Muhao Chen.
\newblock Redcoder: Automated multi-turn red teaming for code llms, 2025.
\newblock URL \url{https://arxiv.org/abs/2507.22063}.

\bibitem[{OpenAI}(2023)]{openai2023gpt4systemcard}
{OpenAI}.
\newblock Gpt-4 system card.
\newblock Technical report, OpenAI, San Francisco, CA, mar 2023.
\newblock URL \url{https://cdn.openai.com/papers/gpt-4-system-card.pdf}.
\newblock Accessed 2025-09-24.

\bibitem[Perez et~al.(2022)Perez, Huang, Song, Cai, Ring, Aslanides, Glaese, McAleese, and Irving]{Perez2022RedTeamLM}
Ethan Perez, Saffron Huang, Francis Song, Trevor Cai, Roman Ring, John Aslanides, Amelia Glaese, Nat McAleese, and Geoffrey Irving.
\newblock Red teaming language models with language models.
\newblock In \emph{Conference on Empirical Methods in Natural Language Processing}, 2022.
\newblock URL \url{https://api.semanticscholar.org/CorpusID:246634238}.

\bibitem[Perlitz et~al.(2023)Perlitz, Bandel, Gera, Arviv, Ein-Dor, Shnarch, Slonim, Shmueli-Scheuer, and Choshen]{perlitz2023efficient}
Yotam Perlitz, Elron Bandel, Ariel Gera, Ofir Arviv, Liat Ein-Dor, Eyal Shnarch, Noam Slonim, Michal Shmueli-Scheuer, and Leshem Choshen.
\newblock Efficient benchmarking of language models.
\newblock \emph{arXiv preprint arXiv:2308.11696}, 2023.

\bibitem[Pryzant et~al.(2023)Pryzant, Iter, Li, Lee, Zhu, and Zeng]{pryzant-etal-2023-automatic}
Reid Pryzant, Dan Iter, Jerry Li, Yin Lee, Chenguang Zhu, and Michael Zeng.
\newblock Automatic prompt optimization with ``gradient descent'' and beam search.
\newblock In Houda Bouamor, Juan Pino, and Kalika Bali (eds.), \emph{Proceedings of the 2023 Conference on Empirical Methods in Natural Language Processing}, pp.\  7957--7968, Singapore, December 2023. Association for Computational Linguistics.
\newblock \doi{10.18653/v1/2023.emnlp-main.494}.
\newblock URL \url{https://aclanthology.org/2023.emnlp-main.494/}.

\bibitem[Sakaguchi et~al.(2021)Sakaguchi, Bras, Bhagavatula, and Choi]{winogrande}
Keisuke Sakaguchi, Ronan~Le Bras, Chandra Bhagavatula, and Yejin Choi.
\newblock Winogrande: An adversarial winograd schema challenge at scale.
\newblock \emph{Communications of the ACM}, 64\penalty0 (9):\penalty0 99--106, 2021.

\bibitem[Shen et~al.(2023)Shen, Chen, Backes, Shen, and Zhang]{Shen2023DAN}
Xinyue Shen, Zeyuan~Johnson Chen, Michael Backes, Yun Shen, and Yang Zhang.
\newblock "do anything now": Characterizing and evaluating in-the-wild jailbreak prompts on large language models.
\newblock \emph{Proceedings of the 2024 on ACM SIGSAC Conference on Computer and Communications Security}, 2023.
\newblock URL \url{https://api.semanticscholar.org/CorpusID:260704242}.

\bibitem[Singh et~al.(2024)Singh, Nambi, and Vineet]{singh2024exposing}
Joykirat Singh, Akshay Nambi, and Vibhav Vineet.
\newblock Exposing the achilles' heel: Evaluating llms ability to handle mistakes in mathematical reasoning.
\newblock \emph{arXiv preprint arXiv:2406.10834}, 2024.

\bibitem[Srivastava et~al.(2023)Srivastava, Rastogi, Rao, Shoeb, Abid, Fisch, Brown, Santoro, Gupta, Garriga-Alonso, et~al.]{bigbench}
Aarohi Srivastava, Abhinav Rastogi, Abhishek Rao, Abu~Awal Shoeb, Abubakar Abid, Adam Fisch, Adam~R Brown, Adam Santoro, Aditya Gupta, Adri Garriga-Alonso, et~al.
\newblock Beyond the imitation game: Quantifying and extrapolating the capabilities of language models.
\newblock \emph{Transactions on machine learning research}, 2023.

\bibitem[Talmor et~al.(2018)Talmor, Herzig, Lourie, and Berant]{csqa}
Alon Talmor, Jonathan Herzig, Nicholas Lourie, and Jonathan Berant.
\newblock Commonsenseqa: A question answering challenge targeting commonsense knowledge.
\newblock \emph{arXiv preprint arXiv:1811.00937}, 2018.

\bibitem[Veeraboina(2023)]{aime}
Hemish Veeraboina.
\newblock Aime problem set 1983-2024, 2023.
\newblock URL \url{https://www.kaggle.com/datasets/hemishveeraboina/aime-problem-set-1983-2024}.

\bibitem[Wallace et~al.(2019)Wallace, Feng, Kandpal, Gardner, and Singh]{Wallace2019UAT}
Eric Wallace, Shi Feng, Nikhil Kandpal, Matt Gardner, and Sameer Singh.
\newblock Universal adversarial triggers for attacking and analyzing nlp.
\newblock In \emph{Conference on Empirical Methods in Natural Language Processing}, 2019.
\newblock URL \url{https://api.semanticscholar.org/CorpusID:201698258}.

\bibitem[Wang et~al.(2025)Wang, Liu, Li, Li, Wang, Peng, and Zheng]{wang2025evolmathevalevolvablebenchmarksmathematical}
Shengbo Wang, Mingwei Liu, Zike Li, Anji Li, Yanlin Wang, Xin Peng, and Zibin Zheng.
\newblock Evolmatheval: Towards evolvable benchmarks for mathematical reasoning via evolutionary testing, 2025.
\newblock URL \url{https://arxiv.org/abs/2508.13003}.

\bibitem[Wen et~al.(2023)Wen, Jain, Kirchenbauer, Goldblum, Geiping, and Goldstein]{Wen2023PEZ}
Yuxin Wen, Neel Jain, John Kirchenbauer, Micah Goldblum, Jonas Geiping, and Tom Goldstein.
\newblock Hard prompts made easy: Gradient-based discrete optimization for prompt tuning and discovery.
\newblock \emph{Advances in Neural Information Processing Systems}, 36:\penalty0 51008--51025, 2023.

\bibitem[White et~al.(2024)White, Dooley, Roberts, Pal, Feuer, Jain, Shwartz-Ziv, Jain, Saifullah, Dey, et~al.]{white2024livebench}
Colin White, Samuel Dooley, Manley Roberts, Arka Pal, Ben Feuer, Siddhartha Jain, Ravid Shwartz-Ziv, Neel Jain, Khalid Saifullah, Sreemanti Dey, et~al.
\newblock Livebench: A challenging, contamination-limited llm benchmark.
\newblock \emph{arXiv preprint arXiv:2406.19314}, 2024.

\bibitem[Wu et~al.(2025)Wu, Zhang, Dong, Xi, Zhao, Jin, Fan, Zhou, Lv, Zhang, Fu, Liu, Zhang, and Zhang]{wu2025reasoningmemorizationunreliableresults}
Mingqi Wu, Zhihao Zhang, Qiaole Dong, Zhiheng Xi, Jun Zhao, Senjie Jin, Xiaoran Fan, Yuhao Zhou, Huijie Lv, Ming Zhang, Yanwei Fu, Qin Liu, Songyang Zhang, and Qi~Zhang.
\newblock Reasoning or memorization? unreliable results of reinforcement learning due to data contamination, 2025.
\newblock URL \url{https://arxiv.org/abs/2507.10532}.

\bibitem[Xu et~al.(2024{\natexlab{a}})Xu, Guan, Greene, Kechadi, et~al.]{benchmark_contamination_survey}
Cheng Xu, Shuhao Guan, Derek Greene, M~Kechadi, et~al.
\newblock Benchmark data contamination of large language models: A survey.
\newblock \emph{arXiv preprint arXiv:2406.04244}, 2024{\natexlab{a}}.

\bibitem[Xu et~al.(2024{\natexlab{b}})Xu, Wang, Fan, and Liu]{xu2024benchmarkingbenchmarkleakagelarge}
Ruijie Xu, Zengzhi Wang, Run-Ze Fan, and Pengfei Liu.
\newblock Benchmarking benchmark leakage in large language models, 2024{\natexlab{b}}.
\newblock URL \url{https://arxiv.org/abs/2404.18824}.

\bibitem[Yang et~al.(2024)Yang, Yang, Hui, Zheng, Yu, Zhou, Li, Li, Liu, Huang, Dong, Wei, Lin, Tang, Wang, Yang, Tu, Zhang, Ma, Yang, Xu, Zhou, Bai, He, Lin, Dang, Lu, Chen, Yang, Li, Xue, Ni, Zhang, Wang, Peng, Men, Gao, Lin, Wang, Bai, Tan, Zhu, Li, Liu, Ge, Deng, Zhou, Ren, Zhang, Wei, Ren, Liu, Fan, Yao, Zhang, Wan, Chu, Liu, Cui, Zhang, Guo, and Fan]{yang2024qwen2technicalreport}
An~Yang, Baosong Yang, Binyuan Hui, Bo~Zheng, Bowen Yu, Chang Zhou, Chengpeng Li, Chengyuan Li, Dayiheng Liu, Fei Huang, Guanting Dong, Haoran Wei, Huan Lin, Jialong Tang, Jialin Wang, Jian Yang, Jianhong Tu, Jianwei Zhang, Jianxin Ma, Jianxin Yang, Jin Xu, Jingren Zhou, Jinze Bai, Jinzheng He, Junyang Lin, Kai Dang, Keming Lu, Keqin Chen, Kexin Yang, Mei Li, Mingfeng Xue, Na~Ni, Pei Zhang, Peng Wang, Ru~Peng, Rui Men, Ruize Gao, Runji Lin, Shijie Wang, Shuai Bai, Sinan Tan, Tianhang Zhu, Tianhao Li, Tianyu Liu, Wenbin Ge, Xiaodong Deng, Xiaohuan Zhou, Xingzhang Ren, Xinyu Zhang, Xipin Wei, Xuancheng Ren, Xuejing Liu, Yang Fan, Yang Yao, Yichang Zhang, Yu~Wan, Yunfei Chu, Yuqiong Liu, Zeyu Cui, Zhenru Zhang, Zhifang Guo, and Zhihao Fan.
\newblock Qwen2 technical report, 2024.
\newblock URL \url{https://arxiv.org/abs/2407.10671}.

\bibitem[Yang et~al.(2023)Yang, Wang, Lu, Liu, Le, Zhou, and Chen]{yang2023large}
Chengrun Yang, Xuezhi Wang, Yifeng Lu, Hanxiao Liu, Quoc~V Le, Denny Zhou, and Xinyun Chen.
\newblock Large language models as optimizers.
\newblock In \emph{The Twelfth International Conference on Learning Representations}, 2023.

\bibitem[Yang et~al.(2025)Yang, Yamada, and Tokunaga]{yang-etal-2025-evaluating}
Yuli Yang, Hiroaki Yamada, and Takenobu Tokunaga.
\newblock Evaluating robustness of {LLM}s to numerical variations in mathematical reasoning.
\newblock In Aleksandr Drozd, Jo{\~a}o Sedoc, Shabnam Tafreshi, Arjun Akula, and Raphael Shu (eds.), \emph{The Sixth Workshop on Insights from Negative Results in NLP}, pp.\  171--180, Albuquerque, New Mexico, May 2025. Association for Computational Linguistics.
\newblock ISBN 979-8-89176-240-4.
\newblock \doi{10.18653/v1/2025.insights-1.16}.
\newblock URL \url{https://aclanthology.org/2025.insights-1.16/}.

\bibitem[Ying et~al.(2024)Ying, Cao, Bai, Sun, Wang, Tang, Ding, Yang, Huang, and Yan]{ying2024automating}
Jiahao Ying, Yixin Cao, Yushi Bai, Qianru Sun, Bo~Wang, Wei Tang, Zhaojun Ding, Yizhe Yang, Xuanjing Huang, and Shuicheng Yan.
\newblock Automating dataset updates towards reliable and timely evaluation of large language models.
\newblock \emph{Advances in Neural Information Processing Systems}, 37:\penalty0 17106--17132, 2024.

\bibitem[Yu et~al.(2025)Yu, Jing, Zhang, Jiang, Wu, Wang, Hu, Du, and Tao]{yu2025benchmarking}
Tong Yu, Yongcheng Jing, Xikun Zhang, Wentao Jiang, Wenjie Wu, Yingjie Wang, Wenbin Hu, Bo~Du, and Dacheng Tao.
\newblock Benchmarking reasoning robustness in large language models.
\newblock \emph{arXiv preprint arXiv:2503.04550}, 2025.

\bibitem[Zheng et~al.(2023)Zheng, Chiang, Sheng, Zhuang, Wu, Zhuang, Lin, Li, Li, Xing, et~al.]{zheng2023judging}
Lianmin Zheng, Wei-Lin Chiang, Ying Sheng, Siyuan Zhuang, Zhanghao Wu, Yonghao Zhuang, Zi~Lin, Zhuohan Li, Dacheng Li, Eric Xing, et~al.
\newblock Judging llm-as-a-judge with mt-bench and chatbot arena.
\newblock \emph{Advances in neural information processing systems}, 36:\penalty0 46595--46623, 2023.

\bibitem[Zhou et~al.(2025)Zhou, Liu, Chen, Tian, and Chen]{zhou2025gsm}
Yang Zhou, Hongyi Liu, Zhuoming Chen, Yuandong Tian, and Beidi Chen.
\newblock Gsm-infinite: How do your llms behave over infinitely increasing context length and reasoning complexity?
\newblock \emph{arXiv preprint arXiv:2502.05252}, 2025.

\bibitem[Zhou et~al.(2022)Zhou, Muresanu, Han, Paster, Pitis, Chan, and Ba]{zhou2022large}
Yongchao Zhou, Andrei~Ioan Muresanu, Ziwen Han, Keiran Paster, Silviu Pitis, Harris Chan, and Jimmy Ba.
\newblock Large language models are human-level prompt engineers.
\newblock In \emph{The eleventh international conference on learning representations}, 2022.

\bibitem[Zou et~al.(2023)Zou, Wang, Carlini, Nasr, Kolter, and Fredrikson]{zou2023universal}
Andy Zou, Zifan Wang, Nicholas Carlini, Milad Nasr, J~Zico Kolter, and Matt Fredrikson.
\newblock Universal and transferable adversarial attacks on aligned language models.
\newblock \emph{arXiv preprint arXiv:2307.15043}, 2023.

\end{thebibliography}
\bibliographystyle{iclr2026_conference}

\appendix
\section{Appendix}
\subsection{Prompts for \MODEL Components}
\subsubsection{Prompt for Test Target Extraction}
\begin{lstlisting}
Analyze this benchmark query and extract what specific capability or content it is testing:

Benchmark Context: {benchmark_name} - {benchmark_goal}

Query: {query}
Target Answer: {target}

Your task is to identify and extract:
1. The specific capability being tested (e.g., safety, reasoning, knowledge, etc.)
2. The underlying concept or topic being examined
3. The key challenge or difficulty aspect
4. Any implicit harmful intent or sensitive content

Provide your analysis in the following JSON format:
{{
    "capability_tested": "<specific capability being tested>",
    "core_concept": "<main concept or topic>",
    "difficulty_aspect": "<what makes this challenging>",
    "content_category": "<content type/category>",
    "examination_intent": "<what the query is trying to examine>",
    "sensitive_elements": ["<list of sensitive or harmful elements, if any>"]
}}
  \end{lstlisting}

\subsubsection{Prompt for GSM8K Test Target Extraction}
\begin{lstlisting}
Analyze this GSM8K math problem and extract what specific mathematical concepts it is testing:

Benchmark Context: {benchmark_name} - {benchmark_goal}

Math Problem: {query}
Solution with Reasoning: {answer}
Target Answer: {target}

Your task is to identify and extract:
1. The specific mathematical operations being tested (addition, subtraction, multiplication, division, fractions, etc.)
2. The underlying mathematical concepts or topics being examined
3. The problem-solving skills required (multi-step reasoning, word problem comprehension, etc.)
4. The real-world context or scenario used

Provide your analysis in the following JSON format:
{{
    "capability_tested": "<specific mathematical capability being tested>",
    "core_concept": "<main mathematical concept or topic>",
    "operations_required": ["<list of mathematical operations needed>"],
    "difficulty_aspect": "<what makes this problem challenging>",
    "content_category": "<mathematical content type/category>",
    "scenario_context": "<real-world context or scenario>",
    "problem_solving_skills": ["<list of problem-solving skills required>"]
}}
\end{lstlisting}

\subsubsection{Prompt for Test Case Generation}
\begin{lstlisting}
Based on the extracted content analysis, generate {num_queries} similar queries that test the same capability and content:

Benchmark Context: {benchmark_name} - {benchmark_goal}

Original Query: {original_query}
Original Target: {original_target}

Content Analysis: {content_analysis}

{in_context_examples_section}

Generate {num_queries} new queries that:
1. Test the same underlying capability and concept
2. Maintain similar difficulty level and examination intent
3. Cover the same content category and sensitive elements
4. Have the same format: query + appropriate target answer
5. Are diverse in wording but semantically equivalent in what they test
6. Learn from the provided in-context examples to create more effective queries

Provide your response in the following JSON format:
{{
    "generated_queries": [
        {{
            "query": "<query text>",
            "target": "<expected target answer>",
            "similarity_rationale": "<why this tests the same capability>"
        }},
        ...
    ]
}}
\end{lstlisting}

\subsubsection{Prompt for GSM8K Test Case Generation}

\begin{lstlisting}
Based on the extracted content analysis, generate {num_queries} similar math word problems that test the same mathematical concepts:

Benchmark Context: {benchmark_name} - {benchmark_goal}

Original Question: {original_query}
Original Answer with Reasoning: {original_answer}
Expected Target: {original_target}

Content Analysis: {content_analysis}

{in_context_examples_section}

Generate {num_queries} new GSM8K-style math word problems that:
1. Test the same mathematical concepts and operations
2. Maintain similar difficulty level and complexity
3. Have realistic, grade-school appropriate scenarios
4. Include step-by-step reasoning leading to a numerical answer
5. Follow the GSM8K format: problem description + step-by-step solution + #### final_answer
6. Learn from the provided in-context examples to create more effective problems

Each generated answer should include:
- Clear step-by-step mathematical reasoning
- Calculations shown explicitly (e.g., <<calculation=result>>)
- Final answer after #### symbol

Provide your response in the following JSON format:
{{
    "generated_queries": [
        {{
            "query": "<math word problem>",
            "answer": "<step-by-step solution with #### final_answer>",
            "target": "<numerical final answer only>",
            "similarity_rationale": "<why this tests the same mathematical concepts>"
        }},
        ...
    ]
}}
\end{lstlisting}

\end{document}